\documentclass[journal]{IEEEtran}
\usepackage{ifpdf}
\ifCLASSINFOpdf
  \usepackage[pdftex]{graphicx}
  \usepackage{graphbox}
  \graphicspath{{figures/}}
  \DeclareGraphicsExtensions{.pdf,.jpeg,.png}
\else
  \usepackage[dvips]{graphicx}
  \graphicspath{{figures/}}
  \DeclareGraphicsExtensions{.eps}
\fi
\usepackage{amsmath}
\usepackage{bm}
\usepackage{amsfonts}
\usepackage[bookmarks=false]{hyperref} 
\usepackage{colortbl}
\usepackage{booktabs}
\usepackage[per-mode=symbol,range-phrase=--,range-units=single]{siunitx}
\usepackage[table,dvipsnames]{xcolor}
\usepackage{cite}
\usepackage{nicefrac}
\usepackage{multirow}
\usepackage{calc}
\usepackage{printlen}
\usepackage{bbding}
\usepackage{amssymb}
\usepackage{pifont}
\usepackage{textcomp}

\usepackage{tikz,xcolor,hyperref}
\definecolor{lime}{HTML}{A6CE39}
\DeclareRobustCommand{\orcidicon}{%
    \begin{tikzpicture}
    \draw[lime, fill=lime] (0,0) 
    circle [radius=0.16] 
    node[white] {{\fontfamily{qag}\selectfont \tiny ID}};    \draw[white, fill=white] (-0.0625,0.095) 
    circle [radius=0.007];    \end{tikzpicture}
    \hspace{-2mm}}
\foreach \x in {A, ..., Z}{%
    \expandafter\xdef\csname orcid\x\endcsname{\noexpand\href{https://orcid.org/\csname orcidauthor\x\endcsname}{\noexpand\orcidicon}}
    }

\usepackage{algorithm}
\usepackage{algorithmic}
\renewcommand{\algorithmicrequire}{\textbf{Input:}}  
\renewcommand{\algorithmicensure}{\textbf{Output:}} 
 
\makeatletter
\newcommand{\removelatexerror}{\let\@latex@error\@gobble}
\makeatother

\usepackage{algorithmic}
\usepackage[normalem]{ulem}
\useunder{\uline}{\ul}{}

\begin{document}
\title{Contrastive Multi-view Subspace Clustering of Hyperspectral Images based on Graph Convolutional Networks}

\author{Renxiang~Guan,\orcidD{}
        Zihao~Li, 
        Xianju~Li,\orcidA{}~\IEEEmembership{Member,~IEEE},
        Chang Tang,\orcidB{}~\IEEEmembership{Member,~IEEE},
        Ruyi Feng\orcidC{}~\IEEEmembership{Member,~IEEE}
        
\thanks{%
    Renxiang Guan, Zihao Li, Xianju Li, Chang Tang and Fuyi Feng are with the Faculty of Computer Science and Key Laboratory of Geological Survey and Evaluation of Ministry of Education, China University of Geosciences, Wuhan 430074, China (e-mail: grx1126@cug.edu.cn; lizihao@cug.edu.cn; ddwhlxj@cug.edu.cn; tangchang@cug.edu.cn; fengry@cug.edu.cn)
}%
\thanks{%
    Renxiang Guan is with College of Computer, National University of Defense Technology, Changsha 410000, China;
}
\thanks{
  A preliminary version of this study was presented at APWEB-WAIM 2023.
}
\thanks{%
  This study was jointly supported by the Natural Science Foundation of China under Grant 42071430, the Opening Fund of Key Laboratory of Geological Survey and Evaluation of Ministry of Education under Grant GLAB2022ZR02 and Grant GLAB2020ZR14.
  (Corresponding authors: Xianju Li)
}%
}
\markboth{SUBMITTED TO IEEE TRANSACTIONS ON GEOSCIENCE AND REMOTE SENSING}%
{Heidler \MakeLowercase{\textit{et al.}}: Monitoring the Antarctic Coastline by Combining Semantic Segmentation and Edge Detection}

\maketitle
\begin{abstract} High-dimensional and complex spectral structures make the clustering of hyperspectral images (HSI) a challenging task. Subspace clustering is an effective approach for addressing this problem. However, current subspace clustering algorithms are primarily designed for a single view and do not fully exploit the spatial or textural  feature information in HSI. In this study, contrastive multi-view subspace clustering of HSI was proposed based on graph convolutional networks. Pixel neighbor textural and spatial-spectral information were sent to construct two graph convolutional subspaces to learn their affinity matrices. To maximize the interaction between different views, a contrastive learning algorithm was introduced to promote the consistency of positive samples and assist the model in extracting robust features. An attention-based fusion module was used to adaptively integrate these affinity matrices, constructing a more discriminative affinity matrix. The model was evaluated using four popular HSI datasets: Indian Pines, Pavia University, Houston, and Xu Zhou. It achieved overall accuracies of 97.61$\%$, 96.69$\%$, 87.21$\%$, and 97.65$\%$, respectively, and significantly outperformed state-of-the-art clustering methods. In conclusion, the proposed model effectively improves the clustering accuracy of HSI.
\end{abstract}

\IEEEpeerreviewmaketitle
\section{Introduction}

\IEEEPARstart{A}{dvancements} in spectral imaging technology have led to the emergence of hyperspectral images (HSI) as potent instruments for detection methodologies, fostering progress in domains such as environmental surveillance \cite{camps2013advances}, geologic investigation \cite{ghamisi2017advances}, national defense and security \cite{eismann2009automated}, and mineral discernment \cite{guan2022classification}. Unlike conventional color imagery, HSI exhibits enhanced resolution and a wealth of spectral data, providing a more precise representation of terrestrial information. This has led to the development of an array of HSI processing techniques.

In recent decades, considerable advancements have been made in supervised HSI classification techniques \cite{chen2022jagan}, \cite{ghamisi2018new}, \cite{qin2018spectral}, encompassing machine learning models like support vector machines \cite{tarabalka2010svm} and deep learning models such as transformer and convolutional neural networks (CNN) \cite{hu2015deep}, \cite{chen2014deep}. These methods are predicated on the necessity of manually labeling data for training samples, a process that is both labor-intensive and reliant on domain expertise \cite{wang2017fast}. To alleviate the human workload and address the limited availability of labeled information in HSI, unsupervised learning approaches such as clustering have garnered significant attention \cite{zhai2021hyperspectral}. Clustering of HSI enables automated data processing and interpretation; however, owing to  the substantial spectral variability and intricate spatial structures of HSI, clustering poses a formidable challenge \cite{borsoi2021spectral}.

HSI clustering entails the classification of pixels into corresponding groups and the segmentation of pixels into distinct categories based on their inherent similarity, maximizing intra-class  and minimizing inter-class similarities \cite{chang2018deep}. Numerous clustering algorithms have been proposed, including those centered on cluster centroids, such as $K$-means \cite{kanungo2002efficient} and fuzzy $c$-means clustering \cite{ghaffarian2014automatic}, as well as those predicated on the spatial density distribution of features, exemplified by the mean shift algorithm \cite{comaniciu2002mean}. and the clustering approach based on ensemble density analysis \cite{chen2017hyperspectral}. Nonetheless, these methods are sensitive to initialization and noise, with an overreliance on similarity measures. To further explore the underlying structure of HSI data, subspace clustering algorithms \cite{vidal2011subspace}, \cite{elhamifar2013sparse}, \cite{hong2020joint} have attracted considerable interest with noteworthy results.

Subspace clustering algorithms effectively integrate traditional feature selection techniques with clustering approaches to process feature subsets or weights corresponding to each data cluster during the partitioning of samples \cite{li2018affinity}. Notable examples include sparse subspace clustering (SSC) \cite{elhamifar2013sparse} and low-rank subspace clustering \cite{vidal2014low}, which rely on identifying the sparse representation matrix of the original data, constructing a similarity graph on the corresponding matrix, and employing spectral clustering to derive clustering outcomes. Recognizing the wealth of spatial environmental information conveyed by the HSIs as a data cube, Zhang et al. \cite{zhang2016spectral} introduced a sparse subspace clustering method to enhance clustering efficacy. In \cite{zhai2016new}, the $l_2$-norm regularized SSC algorithm incorporated adjacent information into the coefficient matrix through l2-norm regularization constraints. Additionally, Huang et al. \cite{huang2019semisupervised} demonstrated the improved utilization of spatial information in localized regions via semi-supervised joint constraints, augmenting the clustering performance of the algorithm.

Although traditional subspace-clustering methods have yielded satisfactory results, their robustness and accuracy remain limited in complex HSI scenarios. Deep clustering models address these shortcomings by extracting profound and robust features \cite{chang2018deep}, \cite{caron2018deep}, \cite{peng2022adaptive}. Lei et al. \cite{lei2020deep} introduced a deep model that employs multilayer autoencoders for self-expression learning. Li et al. \cite{li2021self} integrated self-supervised subspace clustering with an adaptive initialization for HSI and accomplished significant results. To extract robust nonlinear affinity, Cai et al. \cite{cai2021graph} proposed a GR-RSCNet which employs a deep neural network architecture combined with graph regularization for HSI subspace clustering. Moreover, deep clustering models based on graph convolutional networks (GCNs) have gained popularity because of their capacity to capture neighborhood information. Cai et al. \cite{cai2020graph} effectively integrated structural and feature information from a graph representation learning standpoint and proposed a graph convolutional subspace clustering (GCSC) method. Zhang and Huang et al. \cite{zhang2021hypergraph}, \cite{huang2021hybrid} utilized hypergraph convolutional subspace clustering to thoroughly exploit high-order relationships and extensive interdependencies within a HSI.

Despite improvements in the clustering performance, the aforementioned algorithms still have two primary limitations. First, directly applying these methods to HSIs often generates cluster maps with considerable noise due to constraints in discriminative information within the spectral domain, the intricacy of ground objects, and the heterogeneity of spectral features within the same class \cite{hinojosa2021hyperspectral}. Second, these methods are predominantly tested on a single view, and an abundance of experimental evidence suggests that incorporating complementary information from multiple views can effectively enhance clustering accuracy \cite{benediktsson2005classification}, \cite{li2015local}, \cite{jia20173}. Tian et al. \cite{tian2018spatial} conducted multi-view clustering, although it exhibited noise sensitivity. Chen et al. \cite{chen2022tensorial} employed multi-view subspace clustering on a polarized HSI, providing a sparse representation of all data independently by pre-clustering each view. Huang et al. \cite{huang2021hybrid} integrated local and non-local spatial information to discern a shared intrinsic cluster structure, improving the overall clustering performance. Lu et al. \cite{lu2022dynamic} merged spectral and spatial information to establish capability regions using multi-view kernels for collaborative subspace clustering to achieve superior clustering outcomes. The above method shows that multi-view complementary information can improve the accuracy of clustering tasks; however, it is limited to the use of multi-view attributes or multiple branch networks. Not only is there no assistance from other view information in the feature-learning stage, but the information weights of distinct views are neglected, potentially leading to the loss of vital information.

To address these challenges, we introduced  a novel multi-view subspace clustering approach for HSIs that effectively amalgamates textural information with spectral-spatial information. By incorporating a GCN and contrastive learning methodology \cite{cai2022superpixel}, \cite{zhu2021graph}, we aggregated neighborhood information while leveraging the contrastive learning technique to maximize the consistency between views during representation learning. Finally, a more discriminative matrix was obtained using  adaptive fusion affinity matrices based on the attention fusion module. The principal components of this study are as follows.
\begin{enumerate}
    \item We proposed a novel deep multi-view subspace clustering algorithm for HSI clustering, denoted as CMSCGC, capable of concurrently learning textural and spectral-spatial information.
    \item The integration of contrastive learning within the CMSCGC maximized the consistent representation of views, enabling the model to extract more robust feature representations.
    \item An attention-based fusion module was used to adaptively harness the affinity matrices, constructing a more discriminative affinity matrix.
    \item We systematically evaluated our method on four highly competitive datasets, demonstrating that our proposed method significantly outperforms state-of-the-art techniques.
\end{enumerate}

The structure of the rest of this paper is outlined as follows: A brief review of related works on multi-view subspace clustering, HSI clustering, and contrastive learning is provided in Section \ref{sec:related work}. The next section \ref{sec:method} presents the details of the proposed CMSCGC method. The various results are reported and analyzed in Section \ref{sec:Experiments}, followed by a concluding summary in Section \ref{sec:CONCLUSION}.

\section{Related Work}\label{sec:related work}
In this section, we present an overview of multi-view subspace clustering and its foundational models, followed by a concise review of the developments of HSI clustering and contrastive learning.

\subsection{Multi-view Subspace Clustering}
Multi-view clustering approaches \cite{8476207}, \cite{xie2018unifying}, \cite{wang2018multiview} have achieved considerable success across various domains. The field of remote sensing images \cite{huang2021hybrid} can also benefit from multi-view complementary information to facilitate clustering tasks. Given multi-view data $\left\{X^p\right\}_{p=1}^P$, where $\mathrm{X}^p \in \mathrm{R}^{d^p \times n}$ denotes the $p$-th view data with an upper dimension of $d^p$, subspace clustering postulates that every  point can be expressed by a linear combination of other points within the same subspace. Based on this assumption, data $X^p$ for each view were employed as a dictionary to construct the subspace representation model as follows:

\begin{equation}\label{eq:1}
X^p=X^p C^p+E^p
\end{equation}

where $C^p \in \mathbb{R}^{n \times n}$ and $E^p$ represent the self-expression matrix for each view and representation error. Multi-view subspace clustering methods are typically described as follows:

\begin{equation} \label{eq:2}
\begin{gathered}
\min _{S^p}\left\|X^p-X^p C^p\right\|_F^2+\lambda f\left(C^p\right) \\
\text { s.t. } S^p \geq 0, C^{p^T} \mathbf{1}=\mathbf{1}
\end{gathered}
\end{equation}

where $C^{p^T} \mathbf{1}=\mathbf{1}$ ensures that the sum of each column in $C^p$ is 1, indicating that every sample point can be reconstructed through a linear combination of the other samples. $C_{(i,j)}^p$ represents the weight of the edge between the $i$-th and $j$-th samples. Hence, $C^p$ can be viewed as an $n*n$ undirected graph. $f(\cdot)$ is a regularization function, and $\lambda$ is a parameter that balances regularization and loss. Different $f(\cdot)$ functions yield self-expression matrices $C^p$ that satisfy various constraints. For instance, sparse constraints have been applied to matrices \cite{elhamifar2013sparse} or low-rank representations of data \cite{vidal2014low}.

Upon obtaining self-expression matrices $\left\{C^p\right\}_{p=1}^P$ for each view, they are fused to create a unified self-expression matrix $C^p \in \mathbb{R}^{n \times n}$. In certain  methods, this step is performed concurrently with the learning phase across all the views. Using a consistent $C^p$ as the input for spectral clustering yields the final clustering outcome. Because the size of $C^p$ is $n×n$, the time complexity for constructing the self-expression matrix and performing spectral clustering is $O(n^3)$, whereas the space complexity is $O(n^2)$.

\subsection{Hyperspectral Image Clustering}
Owing to the high dimensionality, high correlation, and complex distribution of spectral data, the analysis of hyperspectral data is challenging. Traditional supervised classification methods require a large number of labeled samples that  are often difficult to obtain. In recent years, unsupervised HSI clustering has become an active research topic for overcoming this limitation. HSI clustering groups  pixels with similar spectral features into different classes or clusters without prior data knowledge. Traditional clustering algorithms such as $K$-means are known for their simplicity and efficiency. However, their clustering results are unstable and inaccurate.

Among the existing HSI clustering methods, subspace clustering has shown promising results owing to its robustness. Representative examples include sparse and low-rank subspace clustering \cite{lu2016convex}, \cite{wang2018multiview}. However, these classical algorithms do not fully utilize rich information beyond spectral information. Subsequently, scholars have begun to explore the spatial information in HSIs. For instance, Zhang et al. \cite{zhang2019hyperspectral} integrated spatial-spectral information into non-negative matrix factorization and proposed a robust manifold matrix factorization (RMMF) algorithm with good performance. Cai et al. \cite{cai2020graph} introduced GCNs into  hyperspectral subspace clustering, thereby enhancing the ability to combine clustering models and feature learning.  Liu et al. \cite{liu2022graph} further improved the performance of graph-subspace clustering by incorporating an optimal transport algorithm into the optimization of graph representation learning.

\subsection{Contrastive Learning}
Contrastive learning is an unsupervised technique that learns representations for specific tasks by examining the differences between similar and dissimilar samples. The primary concept is to train the model by comparing the similarities between a pair of samples, mapping them to an embedding space, and comparing their differences in this space by measuring the distances between them \cite{liu2021understand}.

Contrastive learning has been widely applied in various fields, such as computer vision \cite{van2020scan}. For example, the classical SimCLR algorithm \cite{chen2020simple} learns latent representations by maximizing the consistency between different augmented views of the same example, demonstrating the importance of data augmentation in contrastive learning. Similarly, the MoCo algorithm \cite{he2020momentum} approaches contrastive learning from a dictionary lookup perspective, while ensuring a lightweight dictionary and alleviating GPU memory constraints. Gansbeke et al. \cite{van2020scan} generated initial clusters through feature representation learned by pre-training tasks and proposed the SCAN algorithm, which explores the nearest neighbors of each image  and uses this prior knowledge to train the model.

Recently, contrastive learning methods have been introduced to remote sensing image tasks. Hou et al. \cite{hou2021hyperspectral} employed unsupervised contrastive learning to learn robust features from HSI in preparation for subsequent classification tasks. Wang et al. \cite{wang2022reliable} adopted  semisupervised contrastive clustering and selected reliable pseudo-labels for model training to detect changes in remote sensing images. Li et al. \cite{li2022global} proposed a global and local contrastive learning method for semantic segmentation. Wang et al. \cite{wang2023nearest} integrated contrastive learning between HSIs and LiDAR  data for classification tasks.

To the best of our knowledge, existing methods have not introduced multi-view contrastive learning into HSI clustering. Using contrastive learning, the consistency between the same nodes in different views or nodes from the same land-cover category can be maximized, enhancing the feature-learning capabilities and facilitating more robust clustering. The advantages of our method are demonstrated in Section \ref{sec:Experiments}.

\begin{figure*}
\begin{center}
  \includegraphics[width=\linewidth]{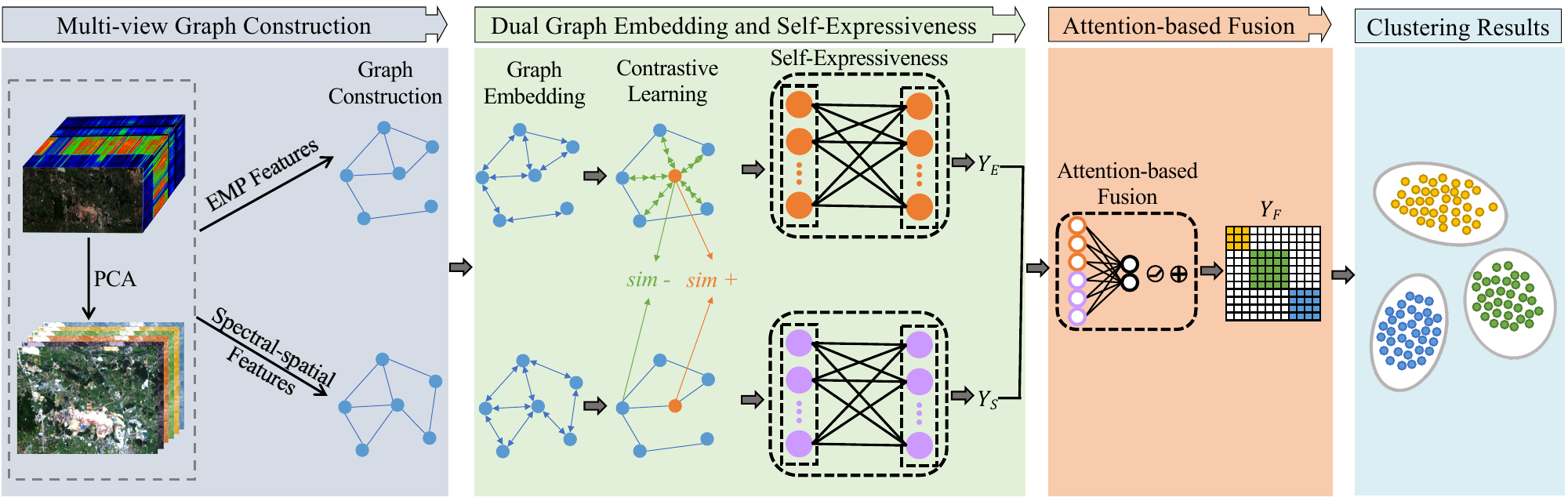}
\end{center}
\caption{Flowchart of our proposed CMSCGC framework. First, we construct multi-view graphs using raw hyperspectral images. Then, a graph convolutional neural network combined with a contrastive learning algorithm is used to learn robust features. An attention-based fusion module is then used to obtain a robust affinity matrix. Finally, spectral clustering or clustering results are applied to the affinity matrix}
\label{fig:main}
\end{figure*}

\section{METHODOLOGY} \label{sec:method}
As shown in Figure \ref{fig:main}, the proposed framework consists of four crucial components:  multi-view graph construction, graph convolutional self-expression, contrastive learning, and attention fusion modules. First, we introduced the multi-view graph construction module, followed by a description of the remaining components.

\subsection{Multi-view Graph Construction}
The key objective of this module is to extract textural and space-spectral features as different views to construct a multi-view graph. However, before this step, a fundamental issue must be addressed: high-dimensional HSIs contain numerous redundant bands. Therefore, we employed principal component analysis  (PCA) \cite{rodarmel2002pca} to reduce the number of bands to dimensions. Subsequently, a sliding window was used to extract the pixel and its adjacent neighboring pixels, with patches representing the data points \cite{cai2020graph}, \cite{fauvel2012advances}. In another branch, the EMP algorithm \cite{benediktsson2005classification} was employed to erode an image while preserving textural information. Therefore, we can obtain multi-view features $\{X^p\}_{p=1}^P$, where $\mathrm{X}^p \in \mathrm{R}^{w \times \mathrm{w} \times d^p \times n}$ and $w$ is the patch size.

Because the GCN can only operate directly on graph-structured data, it is necessary to convert the processed data into a topological graph structure. Multi-view graphs can be defined as $G^p=(V^p,E^p,X^p)$ where $V^p$ and $E^p$ are the corresponding node and edge sets, and the edges are denoted by the adjacency matrix $A^p$. Here, the data in each view are considered nodes of the graph, and the topology is constructed using the $K$-nearest neighbor (KNN) method. In each view, the KNN constructs adjacency matrices $A^p$ by measuring the Euclidean distance between different samples, which describes the global similarity of the data points. The elements in adjacency matrices $A^p$ can be defined as:

\begin{equation} \label{eq3}
A_{i j}^p=\left\{\begin{array}{l}
1, \mathrm{x}_j^p \in \mathcal{N}_k\left(\mathrm{x}_i^p\right) \\
0, \text { otherwise }
\end{array}\right.
\end{equation}

where $x_i^p$ and $x_j^p$ are the columns of $A_ij^p$, and $\mathcal{N}_k\left(\mathrm{x}_i^p\right)$ represents the set of neighbors that include $x_i$ in the $p$-th view. $A_ij^p$ has a value of either {0,1}, depending on whether $x_i^p$ and $x_j^p$ are similar.

\subsection{Graph Convolutional Self-expression Module}
In HSI clustering tasks, learning the topological relationships between objects has long been overlooked and conventional CNN struggle to learn these relationships effectively. Thus, introducing models that can enhance object topology learning could potentially improve clustering accuracy.

Unlike models based on CNN, GCN can fully exploit the dependency relationships between nodes and the feature information of each neighboring node through graph convolution operations. Mathematically, the formula for the spectral graph convolution at the $l$-th layer is

\begin{equation} \label{eq4}
\begin{aligned}
H^p(l+1) & =f\left(H^p(l), \widetilde{\boldsymbol{A}}^p ; W^{(l)}\right) \\
& =\sigma\left(\widetilde{\boldsymbol{D}}^{p^{-\frac{1}{2}}} \widetilde{\boldsymbol{A}}^p \widetilde{\boldsymbol{D}}^{p^{-\frac{1}{2}}} H^p(l) W^p(l)\right).
\end{aligned}
\end{equation}

In this equation, $\widetilde{A}^p = A^p + I_N$, where $I_N$ is the identity matrix; $\widetilde{D}^p$ and $W^p (l)$ represent the degree matrix of nodes in each view and the weight matrix of the $(l+1)$-th layer, respectively; $\sigma$ denotes the activation function; $H(l)$ is for the data representation at layer $l$, and when $l$ equals 0, $H^p (0)$ represents the input multi-view data.

Inspired by the definition of graph convolutional subspace clustering in [31], we modified the traditional self-expression using the self-expression coefficient matrix to obtain more robust information as follows:

\begin{equation}\label{eq5}
\begin{aligned}
\mathrm{X}^p & =\mathrm{X}^p \bar{A}^p C^p \\
& =Z^p C^p \text { s.t. } \operatorname{diag}(C)=0.
\end{aligned}
\end{equation}

where $C^p$ is the self-expression coefficient matrix for each view, and $X\mathrm{A}C$ can be regarded as a special linear graph convolution operation parameterized by $C$, where $\mathrm{A}$ represents the normalized adjacency matrix. The final graph self-expression is defined as

\begin{equation}\label{eq6}
\min_{C^p}\frac12(\lambda\parallel Z^pC^p-X^p\parallel_q+\lambda\parallel C^p\parallel_q)\text{ s.t. diag }(C^p)=0,
\end{equation}

where $q$ denotes an appropriate matrix norm, and $\lambda$ is the balancing coefficient. The purpose of graph convolutional self-expression is to reconstruct the original data using a self-expressive dictionary matrix $Z$. As $Z$ includes the global structural information, a clearer dictionary can be obtained, generating a robust affinity matrix. By setting $\|\cdot\|_q$ as the Frobenius norm, the self-expression loss function can be obtained as follows:

\begin{equation}\label{eq7}
\begin{aligned}
\mathcal{L}_{\mathcal{S}E}& =\frac12(\parallel Z^pC^p-X^p\parallel_F+\lambda\parallel C^p\parallel_F)  \\
&=\frac12tr\left[(Z^pC^p-X^p)^T(Z^pC^p-X^p)+\lambda C^{pT}C^p\right]
\end{aligned}
\end{equation}

The obtained multi-view self-expression coefficient matrices $C^p$ can be used to construct affinity matrices $Y^p$:

\begin{equation}\label{eq8}
Y^p=\frac12(|C^p|+|C^p|^\mathrm{T})
\end{equation}

\subsection{Contrastive Learning}
The effective utilization of information from multiple views is a significant challenge faced by multi-view clustering. That is, we must learn consistent information from different perspectives to promote better clustering. The aforementioned GCN has already extracted deep multi-view features. To facilitate the subsequent steps, we represent the features of the nodes as $z$.

Recently, contrastive learning has demonstrated excellent performance in clustering tasks because it can maintain the desired invariance of data by constraining the distribution of similarities and dissimilarities. The concept of contrastive learning is simple: construct positive and negative samples, minimize the distance between positive samples, and remove negative samples to learn better node representations. A significant amount of work has been conducted  on constructing better positive and negative samples, which play an important role in subsequent tasks. In this study, there were naturally high-quality positive samples, that is, nodes of the same graph in different views, and random nodes from different graphs were selected as negative samples.

For any node $i$ on the graph, we consider its node representation $z_i$ in one view as the anchor point, and the corresponding node representations in the other views, denoted as $z_{i}^{\prime}$, form the positive samples. As mentioned earlier, we want $z_i$ to be as far  from the negative samples as possible and closer to the positive samples. The loss function is designed as follows:

\begin{equation} \label{eq9} 
\ell(z_{i},z_{i}^{\prime})=-\log\frac{e^{(sim{(z_{i}z_{i}^{\prime})}/\tau)}}{\sum_{k=1}^{N}e^{(sim{(z_{i}z_{k}^{\prime})}/\tau)}+\sum_{k=1}^{N}e^{(sim{(z_{i}z_{k})}/\tau)}}
\end{equation}

where $\tau$ represents the temperature parameter, $sim(\cdot)$ is the cosine similarity function, $z_i^{\prime}$ represents the negative sample representation in other views, and $z_i$ represents the representation of the negative pair within the same layer. The overall objective for minimizing is the average of Equation (9) for all given positive pairs, which is given by the following expression:

\begin{equation} \label{eq10}
\mathcal{L}_{con}=\frac1{2N}\sum_{i=1}^N[\ell(z_i,z_i^{\prime})+\ell(z_i^{\prime},z_i)]
\end{equation}

\subsection{Attention Fusion Module}
After obtaining affinity matrices $Y^p$ from Eq.\ref{eq8}, we must fuse them to construct the final affinity graph and apply spectral clustering to the final affinity matrix. We utilized an attention-based fusion module to learn the importance $a^p$ of each view as follows:

\begin{equation} \label{eq11}
a^p=att(Y^p)
\end{equation}

where $a^p\in R^{n\times1}$ are used to measure the importance of each view. Details of the attention module are provided below. In the first step, we concatenate affinity matrices $Y^p$ as $[Y_1\cdots Y_p]\in\mathbb{R}^{n \times pn}$ and introduce a weight matrix $S\in\mathbb{R}^{pn \times p}$ to capture the relationships between the self-expression matrices. Next, we apply the tanh function to the product of $[Y_1\cdots Y_p]$ and $W$ for a nonlinear transformation. Finally, we use softmax and the $l_2$ function to normalize the attention values, resulting in the final weight matrix

\begin{equation} \label{eq12}
a^p=\ell_2(softmax(tanh([Y_1\cdots Y_p]\cdot S)))
\end{equation}

The weight matrix can be obtained to realize the fusion operation, and fused self-expression matrix $Y_F$ is

\begin{equation} \label{eq13}
Y_F=f_\cdot(Y_1\cdots Y_p)=\sum_{i=1}^N(a_i1)\odot Y_i
\end{equation}

where $1\in\mathbb{R}^{1\times n}$ is a matrix with all elements being 1, and $\odot$ represents the Hadamard product of the matrix.

\begin{figure}[!t]
		\label{alg:CMSCGC}
		\renewcommand{\algorithmicrequire}{\textbf{Input:}}
		\renewcommand{\algorithmicensure}{\textbf{Output:}}
		\removelatexerror
		\begin{algorithm}[H]
			\caption{Learning Procedure for CMSCGC}
			\begin{algorithmic}[1]
				\REQUIRE HSI data $\mathbf {X}$, the input size $w$, the number of neighbors: $k$, the regularization coefficient:$\lambda$, ant the training epochs $T$          
				\ENSURE Clustering results  
				\STATE Initialize the parameters of the network.
				\WHILE { $epoch \leq T$ }
				\STATE Feed $\mathbf {X}$ into Multi-view Construction module to get the multi-view data $\{X^p\}_{p=1}^P$.
                    \STATE Construct graph using Eq.(\ref{eq3})
                    \STATE Using graph convolution by to get the features $Z^p$ Eq.(\ref{eq4})
                    \STATE Using multi-view contrastive learning to get the deep features by Eq.(\ref{eq9})
                    \STATE Undate parameters of the network by miniziming Eq.(\ref{eq10})
				\ENDWHILE
				\STATE Perform graph convolutional subspace clustering process to get the cluster labels.
			\end{algorithmic}
		\end{algorithm}
	\end{figure}

\section{EXPERIMENTS}\label{sec:Experiments}
We demonstrate the clustering effectiveness of the proposed model on four benchmark datasets and compare them with several state-of-the-art clustering models. Some details about the experimental setup and results are presented in later sections.

\begin{table}[!t] \label{tb:datasets}
\caption{STATISTICS OF THE DATASETS USED IN OUR EXPERIMENTS}
\centering
\resizebox{\columnwidth}{!}{
\begin{tabular}{ccccc}
\hline
\textbf{Datesets} & \textbf{Indian Pines} & \textbf{Pavia University} & \textbf{Houston} & \textbf{Xu Zhou} \\
\hline
\textbf{Pixels}   & 85×70 & 200×100 & 349×680 & 100×260 \\
\textbf{Abscissa} & 30-115 & 150-350 & 0-349  & 0-100 \\
\textbf{Ordinate} & 24-94 & 100-200 & 0-680 & 0-260 \\
\textbf{Channels} & 200   & 103     & 144     & 436     \\
\textbf{Samples}  & 4391  & 6445    & 6048    & 12043   \\
\textbf{Clusters} & 4     & 8       & 12      & 5      \\
\hline
\end{tabular}
}
\end{table}

\begin{table}[] 
\label{tb:parameters}
\caption{IMPORTANT HYPERPARAMETER SETTINGS IN OUR EXPERIMENTS}
\begin{tabular}{ccccc}
\hline
\textbf{Datesets} & \textbf{Indian Pines} & \textbf{Pavia University} & \textbf{Houston} & \textbf{Xu Zhou} \\
\hline
\textbf{$w$}  & 11  & 11   & 11   & 7   \\
\textbf{$k$}  & 25  & 30   & 25   & 35  \\
\textbf{$\lambda$} & 100 & 1000 & 1000 & 100 \\
\hline
\end{tabular}

\end{table}

\subsection{Setup}
\subsubsection{Dataset}
To provide a fair assessment of the model's effectiveness, it is crucial to employ a diverse dataset. This paper evaluates the model's classification prowess using three widely recognized HSI datasets: Indian Pines, Pavia University, Houston-2013, and Xu Zhou, which were acquired using AVIRIS, ROSIS, ITRES CASI-1500, and HySpex sensors. For computational efficiency, we selected a subset of the scenes for our experiments. Details of the selected sub-scenes are listed in Table \ref{tb:datasets}.

\textbf{Indian Pines:} This HSI was acquired using the American AVIRIS sensor in the Indian Pines. The selected sub-scene is located at [30–115, 24–94], incorporating 200 spectral bands per pixel. The spatial resolution of the dataset stands at 20 meters, encompassing four distinct labels.

\textbf{Pavia University:} The HSI was acquired by the German ROSIS sensor in 2003, it had a spatial resolution of 1.3 meters and a spectral range of 0.43–0.86, divided into 115 bands. After removing the 12 noisy bands, the remaining 103 bands were used for clustering. The selected sub-scene was located at [150–350, 100–200].

\textbf{Houston 2013:}  It was collected by the ITRES CASI-1500 sensor in Texas, USA, and its adjacent rural areas in 2013. After deleting the noise bands, the remaining 144 valid spectral bands are used for experiments. The Houston dataset has a spatial resolution of 2.5mpp, and contains 12 land cover categories, such as tree, soil, water, healthy grass, running track, tennis court, etc. The selected sub-scene was located at [0–349, 0–680].

\textbf{Xu Zhou:} This HSI was located in the suburban mining area of Xuzhou City, Jiangsu Province, China. The data were obtained using an airborne HySpex imaging spectrometer with a high resolution of 0.73 meters. Additionally, this scene contains data for 436 bands with a spectral range of 415–2508 nm. The selected sub-scene was located at [0–100, 0–260] and contains 5 land cover categories including crops, bareland, red-tiles, trees, and cement.

\subsubsection{Evaluation Metrics}
To quantitatively analyze the advantages and disadvantages of the algorithm constructed in this article and the comparison algorithm, we use three commonly used evaluation metrics to evaluate the model proposed in this article, they are clustering accuracy (ACC), normalized mutual information (NMI), and Kappa coefficient (Kappa). The following section introduces the evaluation metrics in detail.

(1) ACC is one of the most widely used clustering evaluation metrics. This refers to the proportion of correctly predicted samples among all samples. The formula for calculating the ACC is as follows:

\begin{equation} \label{eq14}
ACC=\frac{\sum_{1}^{N}\delta(y_{i},\mathrm{map}(c_{i}))}{N}
\end{equation}

where $N$ is the total number of samples, $y_i$ represents the true label of data sample $x_i$, and $c_i$ represents the predicted label of the algorithm. $\delta(\cdot)$ represents the indicator function, and its value is 1 if $y_i=map(c_i)$; otherwise, its value is 0. $map(\cdot)$ represents the mapping function, which rearranges the obtained labels to create the best match of the true labels.

(2) NMI is an index used to measure the similarity between two clustering results:

\begin{equation} \label{eq15}
NMI=\frac{I(y_i,c_i)}{\sqrt{H(y_i)H(c_i)}}
\end{equation}

where $I(y_i, c_i)$ represents the mutual information between true and predicted labels. The larger value of $I(y_i, c_i)$ represents the closer the relationship between $y_i$ and $c_i$. $H(\cdot)$ represents information entropy, which is the average amount of information corresponding to the entire probability distribution. The denominator in the formula is used to normalize the mutual information within the range [0,1].

(3) Kappa is defined as:

\begin{equation} \label{eq16}
\text { Kappa }=\frac{N \sum_{\mathrm{i}=1}^m h_{i i}-\sum_{\mathrm{i}=1}^m\left(h_{i+} h_{+i}\right)}{N^2-\sum_{\mathrm{i}=1}^m\left(h_{i+} h_{+i}\right)},
\end{equation}

where $h_{i+}$ represents the total number of samples of the $i$-th category, and $h_{+i}$ represents the number of samples classified as $i$-th. $N$ is the total number of class samples.

\subsection{Baseline}
In our experiment, we compared our proposed model with ten benchmark methods, including two classic clustering algorithms, SC and SSC, as well as several outstanding clustering algorithms developed in recent years:

\begin{itemize}
\item \textbf{K-means} \cite{kanungo2002efficient} is an outstanding representative of traditional unsupervised learning.

\item \textbf{SSC} \cite{zhang2016spectral} maintains subspace data associations by imposing sparse regularization.

\item \textbf{$l_2$-SSC} \cite{zhai2016new} combines the $l_2$-norm and considers more extensive spatial constraints.

\item \textbf{EGCSC} \cite{cai2020graph} first introduces graph convolutional subspace clustering for HSI.

\item \textbf{RMMF} \cite{zhang2019hyperspectral} integrates spectral-spatial information into non-negative matrix factorization for affinity matrix learning.

\item \textbf{GCOT} \cite{liu2022graph} introduces optimal transport into graph representation learning, facilitating the subsequent learning of natural affinity matrices.

\item \textbf{GR-RSCNet} \cite{cai2021graph} employs a deep neural network architecture combined with graph regularization for HSI subspace clustering.

\item \textbf{HMSC} \cite{huang2021hybrid} constructs multi-view data and integrates local and nonlocal spatial information from each view.

\item \textbf{HGCSC} \cite{zhang2021hypergraph} introduces a multihop hypergraph convolution, representing intraclass relationships as edges in the hypergraph.
\end{itemize}

\begin{table*}[!t] \label{tb:main}
\caption{CLUSTERING RESULTS (OA/NMI/KAPPA) OF VARIOUS METHODS ON FIVE BENCHMARK DATASETS. THE BEST AND SUBOPTIMAL RESULTS ARE SHOWN IN BOLD AND UNDERLINED, RESPECTIVELY.}
\begin{tabular}{c|cccccccccccc}
\hline
\textbf{Dataset} &
  \textbf{Metric} &
  \textbf{$k$-means} &
  \textbf{SSC} &
  \textbf{$l_2$-SSC} &
  \textbf{RMMF} &
  \textbf{NMFAML} &
  \textbf{EGCSC} &
  \textbf{GCOT} &
  \textbf{HMSC} &
  \textbf{HGCSC} &
  \multicolumn{1}{l}{\textbf{GR-RSCNet}} &
  \textbf{CMSCGC} \\
\hline
\multirow{3}{*}{\textbf{InP.}} & OA    & 0.5953 & 0.6558 & 0.6645 & 0.7121 & 0.8508 & 0.8483 & 0.8404 & 0.8901 & 0.9244       & {\ul 0.9317}    & \textbf{0.9761} \\
                               & NMI   & 0.4483 & 0.6025 & 0.5382  & 0.4985 & 0.7264 & 0.6442 & 0.7654 & 0.86   & {\ul 0.8921} & 0.8241          & \textbf{0.9234} \\
                               & Kappa & 0.4404 & 0.5528 & 0.5263  & 0.5609 & 0.7809 & 0.6422 & 0.6258 & 0.9    & 0.8307       & {\ul 0.9008}    & \textbf{0.9657} \\
\hline
\multirow{3}{*}{\textbf{PaU.}} & OA    & 0.6027 & 0.6548 & 0.691  & 0.7704 & 0.8967 & 0.8442 & 0.9102 & 0.9482 & 0.9531       & \textbf{0.9773} & {\ul 0.9669}    \\
                               & NMI   & 0.5306 & 0.6706 & 0.6251 & 0.7388 & 0.9216 & 0.8401 & 0.8804 & 0.92   & 0.9383       & {\ul 0.9689}    & \textbf{0.9702} \\
                               & Kappa & 0.6791 & 0.5736 & 0.6881 & 0.6804 & 0.8625 & 0.7968 & 0.9233 & 0.83   & 0.9440        & \textbf{0.9702} & {\ul 0.9565}    \\
\hline
\multirow{3}{*}{\textbf{Hou.}} & OA    & 0.3932 & 0.4581 & 0.5097 & 0.7309 & 0.6346 & 0.6238 & —      & —      & 0.7315       & {\ul 0.8399}    & \textbf{0.8721} \\
                               & NMI   & 0.4101 & 0.4961 & 0.5816 & 0.7797 & 0.7959 & 0.7754 & —      & —      & 0.7004       & {\ul 0.8731}    & \textbf{0.9168} \\
                               & Kappa & 0.3289 & 0.4013 & 0.4811 & 0.6111 & 0.591  & 0.5812 & —      & —      & {\ul 0.8309} & 0.8221          & \textbf{0.858}  \\
\hline
\multirow{3}{*}{\textbf{XuZ}}  & OA    & 0.5778 & 0.6638 & 0.6838 & 0.6111 & 0.8178 & 0.7834 & —      & —      & 0.9157       & {\ul 0.9192}    & \textbf{0.9765} \\
                               & NMI   & 0.4049 & 0.5357 & 0.5969 & 0.6253 & 0.7919 & 0.6975 & —      & —      & {\ul 0.8862} & 0.8739          & \textbf{0.9272} \\
                               & Kappa & 0.4175 & 0.4767 & 0.6592 & 0.4361 & 0.7459 & 0.6470  & —      & —      & 0.8470        & {\ul 0.8649}    & \textbf{0.9681} \\
\hline
\end{tabular}
\end{table*}

\begin{figure*}
\begin{center}
  \includegraphics[width=\linewidth]{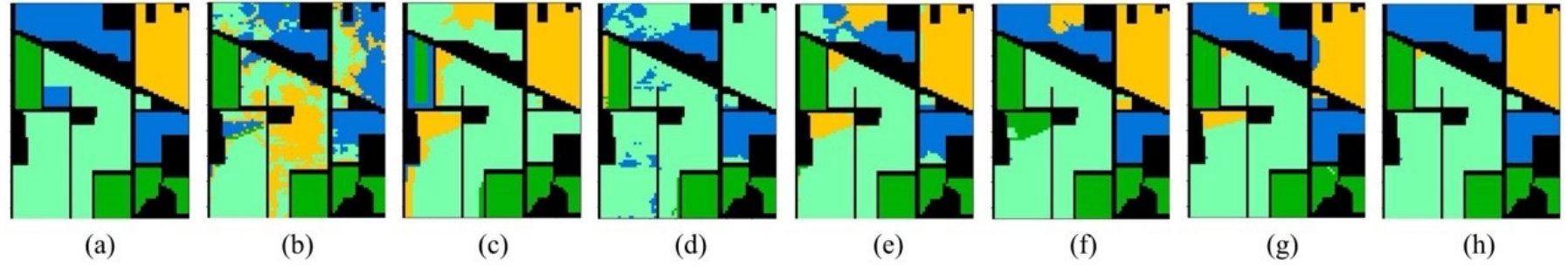}
\end{center}
\caption{Cluster visualization on the Indian Pines dataset: (a) ground truth, (b) $k$-means, (c) SSC, (d) $l_2$-SSC, (e) EGCSC, (f) HGCSC, (g) GR-RSCNet, and (h) our proposed CMSCGC.
}
\label{fig:indian}
\end{figure*}

\begin{figure*}
\begin{center}
  \includegraphics[width=\linewidth]{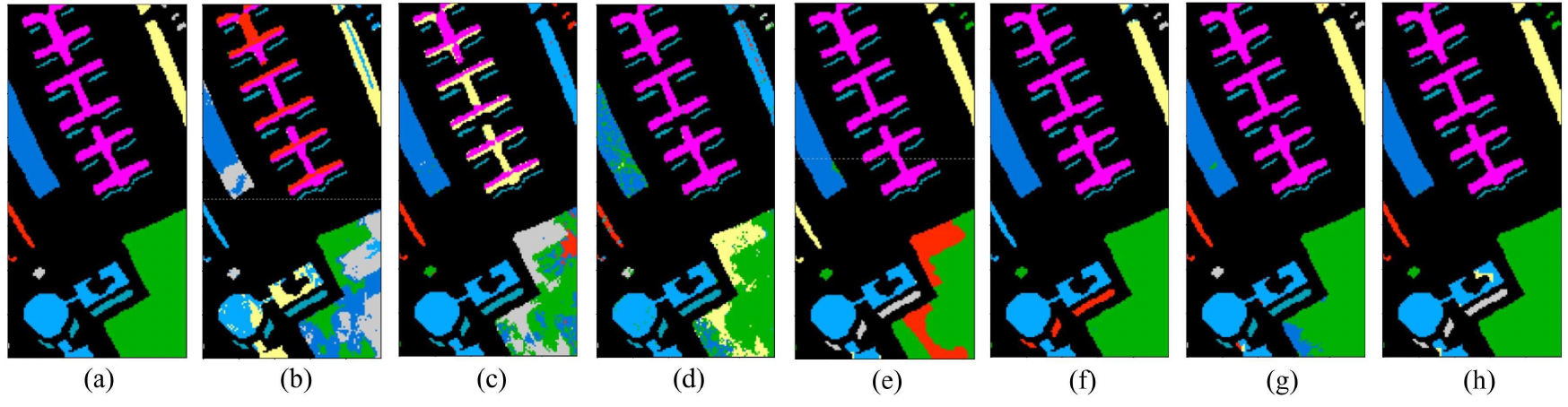}
\end{center}
\caption{Cluster visualization on the Pavia University dataset: (a) ground truth, (b) $k$-means, (c) SSC, (d) $l_2$-SSC, (e) EGCSC, (f) HGCSC, (g) GR-RSCNet, and (h) our proposed CMSCGC.
}
\label{fig:PaviaU}
\end{figure*}

\begin{figure*}
\begin{center}
  \includegraphics[width=\linewidth]{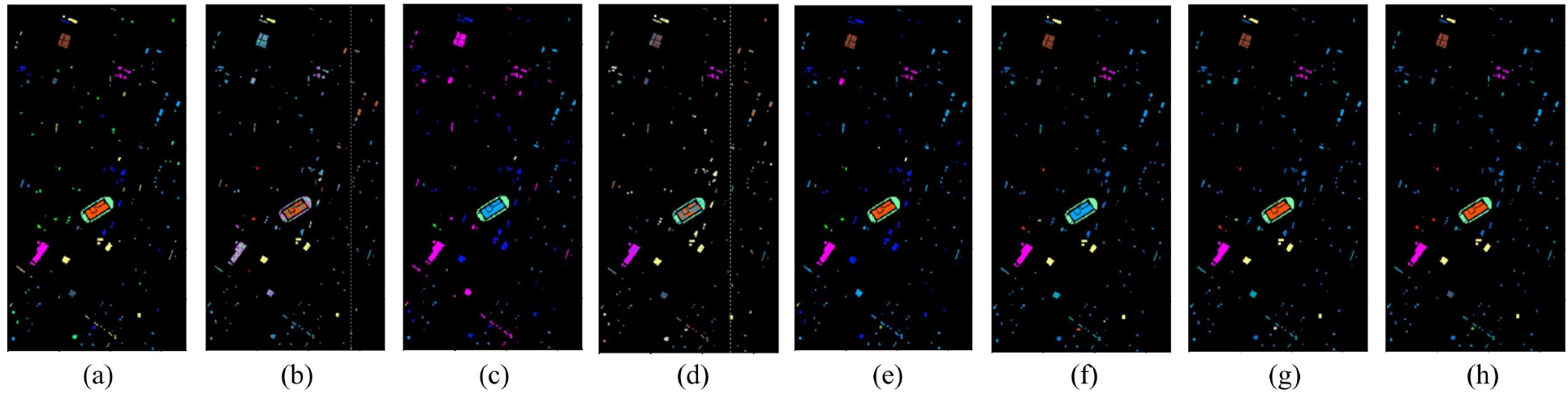}
\end{center}
\caption{Cluster visualization on the Houston-2013 dataset: (a) ground truth, (b) $k$-means, (c) SSC, (d) $l_2$-SSC, (e) EGCSC, (f) HGCSC, (g) GR-RSCNet, and (h) our proposed CMSCGC.
}
\label{fig:houston}
\end{figure*}

\begin{figure*}
\begin{center}
  \includegraphics[width=\linewidth]{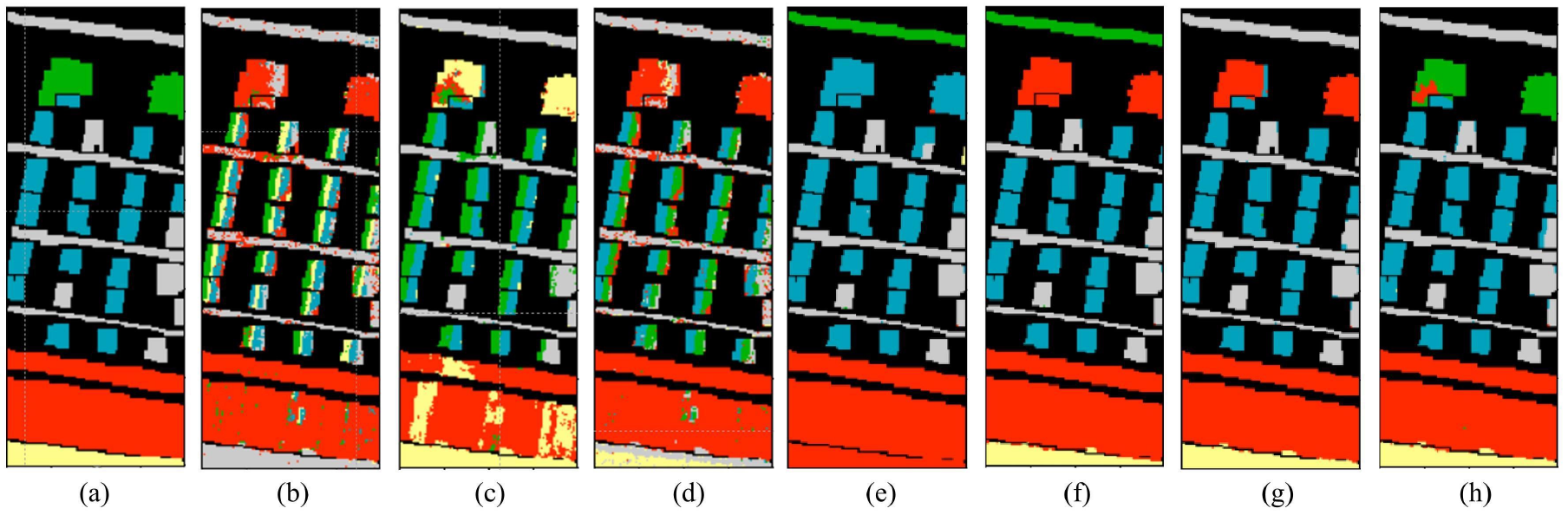}
\end{center}
\caption{Cluster visualization on the Xu Zhou dataset: (a) ground truth, (b) $k$-means, (c) SSC, (d) $l_2$-SSC, (e) EGCSC, (f) HGCSC, (g) GR-RSCNet, and (h) our proposed CMSCGC.
}
\label{fig:xuzhou}
\end{figure*}

\begin{figure}
\begin{center}
  \includegraphics[width=\linewidth]{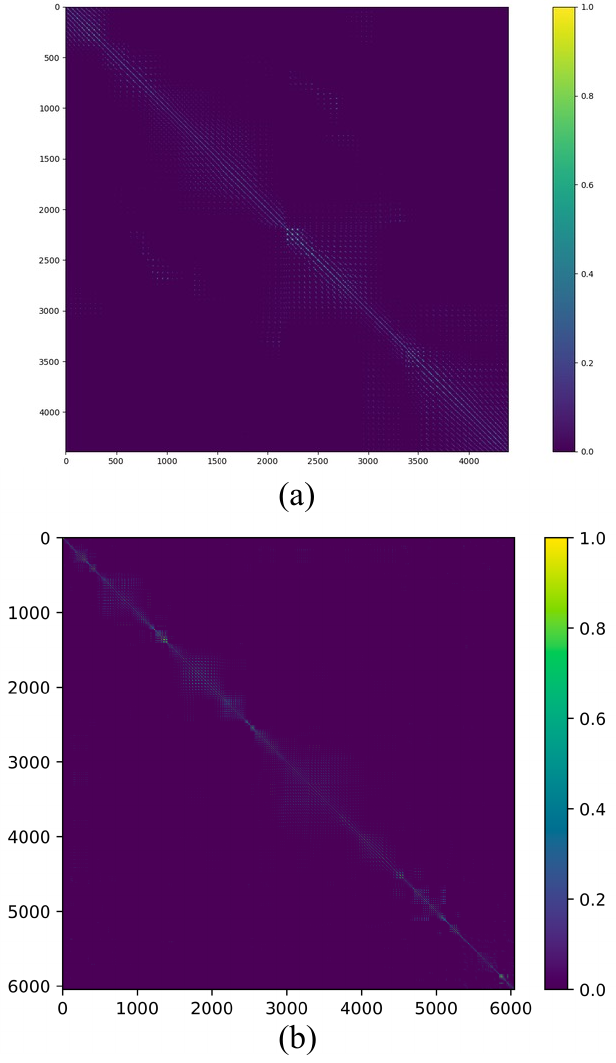}
\end{center}
\caption{Visualization of the affinity matrix learn on (a) Indian Pines. (b) Houston.
}
\label{fig:affinity}
\end{figure}

\subsection{Quantitative Results}

Table \ref{tb:main} shows the clustering performance of the CMSCGC and other models for the four datasets. Our model achieved the best clustering performance in most cases across the three evaluation metrics, OA, NMI, and Kappa, significantly outperforming the other methods. Specifically, for the Indian Pines, Houston, and Xu Zhou datasets, the proposed method achieved a 3–5$\%$ improvement in accuracy compared with  the second-best approach. Furthermore, we observed the following trends.

(1) The introduction of deep learning methods and regularization in clustering effectively improves the accuracy. Deep-learning-based approaches, such as EGCSC and GR-RSCNet, show significant improvements compared to traditional models, indicating that nonlinear projections help extract more suitable representations for clustering complex real-world datasets. In traditional models, $l_2$-SSC introduces $l_2$-norm regularization, improving the accuracy by 10$\%$ on the Indian Pines  dataset compared with the SSC method. The EGCSC introduces graph regularization, and our model introduces contrastive learning regularization, which also achieves significant improvements.

(2) The introduction of complementary information from multiple views improved the clustering accuracy. In addition to our model, NMFAML  is another method in the comparison experiments that extracts additional feature information by incorporating homogeneous information. Similarly, our model combines textural information and uses a GCN to aggregate neighborhood information and effectively improve clustering accuracy, achieving 97.61$\%$, 96.69$\%$, 87.21$\%$, and 97.65$\%$ accuracy on the Indian Pines, Pavia University, Houston2013, and Xu Zhou datasets, respectively.

(3) CMSCGC outperformed EGCSC in terms of the experimental accuracy for all four datasets. Specifically, our model improved by 12.48$\%$, 11.87$\%$, 24.83$\%$, and 19.31$\%$ respectively for all datasets. This indicates that the introduction of multi-view information, contrastive learning regularization, and attention fusion modules can effectively enhance clustering accuracy. Our model provides a new approach for HSI clustering.

\subsection{Qualitative Comparison of Different Methods}
Figures \ref{fig:indian}-\ref{fig:xuzhou} show the visualization results of various clustering methods on the Indian Pines, Pavia University, Houston2013, and Xu Zhou datasets, respectively. Section (a) of each image shows the ground truth of the corresponding dataset with the irrelevant background removed. As spectral clustering is an unsupervised method, the colors representing the same class in different methods may vary; however, this does not affect our visualization results. Sparse subspace clustering algorithms can extract low-dimensional information from high-dimensional data structures but do not consider spatial constraints. Moreover, traditional subspace clustering methods struggle to accurately model the HSI structure, making them  inferior to deep-learning-based methods.

Our model achieved the best visualization results for all the datasets. Owing to the adaptive aggregation of textural information, the CMSCGC has the least salt-and-pepper noise in the classification images compared to the other methods. Simultaneously, because the GCN model can aggregate neighboring node information, it maintains improved  homogeneity within the same class of regions, and the image boundaries remain relatively intact.

\begin{figure}
\begin{center}
  \includegraphics[width=\linewidth]{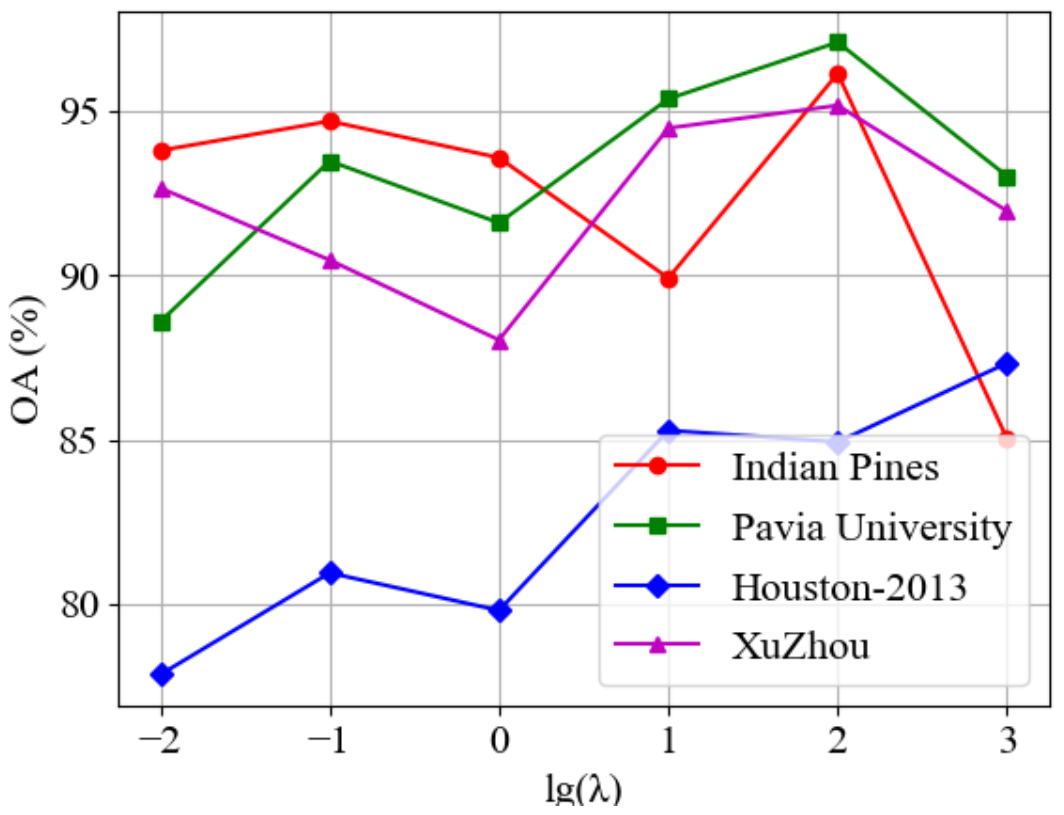}
\end{center}
\caption{Performance comparison on four datasets with different regularization coefficient $\lambda$
}
\label{fig:lambda}
\end{figure}

\subsection{Parameter Analysis}
In this study, we investigated the impacts of the three most important hyperparameters on the CMSCGC framework. Table \ref{tb:parameters} provides the set of suitable parameter settings used in the experiments.

\textbf{Sensitivity to regularization coefficient $\lambda$:}  We chose $\lambda$ from the set $S = {0.01, 0.1, 1, 10, 100, 1000}$, and Figure \ref{fig:lambda} shows the performance of CMSCGC. On the Houston dataset, relatively smaller $\lambda$ values result in lower ACC, NMI, and Kappa values. This is possibly because the Houston dataset contains dispersed similar data; therefore, when performing SE, the regularization term for learning the structural affinity matrix is crucial for maintaining connectivity within the same subspace. On the Xu Zhou dataset, our method can achieve nearly optimal performance in terms of the three metrics across a wide range of $\lambda$ values, indicating that our method has certain robustness to $\lambda$.\

\begin{figure*}
\begin{center}
  \includegraphics[width=\linewidth]{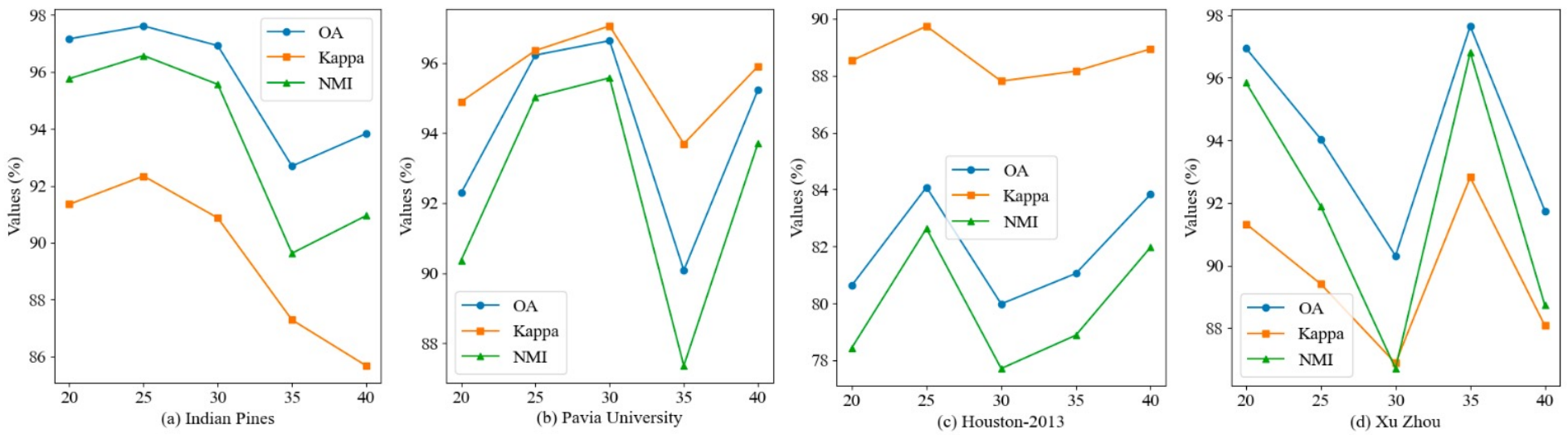}
\end{center}
\caption{Performance comparison on four datasets with different number of neighbors $k$
}
\label{fig:k}
\end{figure*}

\textbf{Sensitivity to nearest neighbors $k$:} Nearest neighbors $k$ are used to construct the graph embedding of the HSI dataset. To investigate the impact of the top $k$ neighbors in the $k$-NN graph, we studied the influence of $k$ while keeping other parameters fixed. The value of $k$ ranges from 20 to 40, with increments of 5. As shown in Figure \ref{fig:k}, if $k$ becomes too large, all three metrics of the model decrease. This is because an excessively large $k$ can lead to the over smoothing problem. Therefore, we set the $k$ values for the four datasets to 25, 30, 25, and 35, respectively.

\begin{figure*}
\begin{center}
  \includegraphics[width=\linewidth]{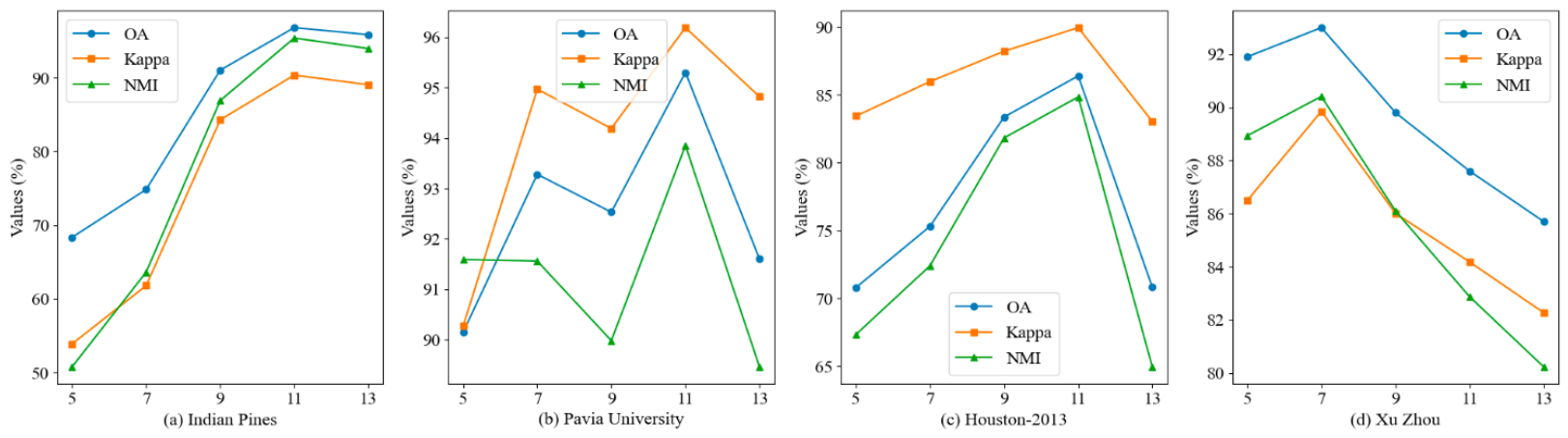}
\end{center}
\caption{Performance comparison on four datasets with different input size $w$
}
\label{fig:inputsize}
\end{figure*}

\textbf{Sensitivity to patch window size $w$:} Window size $w$ is the size of the spatial window used to extract spatial information. We conducted experiments with fixed other parameters to explore the influence of $w$. The value of w starts at 5, with increments of 2, until reaching 13. As shown in Figure \ref{fig:inputsize},  if $w$ is too small, the patches may contain more redundant spatial information, resulting in a loss of accuracy. Conversely , the model may not be able to extract sufficient spatial information, which would also reduce its performance. When parameter $w$ is set to 11, 11, 11, and 9, respectively, CMSCGC achieves the best results for the four datasets.

\begin{table*}[] \label{tb:ablation}
\caption{ABLATION STUDIES (OA) FOR DIFFERENT VIEWS, CONTRACTIVE LEARNING, AND ATTENTION FUSION MODULE FOR CMSCGC. THE BEST AND SUBOPTIMAL RESULTS ARE SHOWN IN BOLD AND UNDERLINE, RESPECTIVELY}
\resizebox{\textwidth}{!}{
\begin{tabular}{ccccccccc}
\hline
No. &
  \begin{tabular}[c]{@{}c@{}}Texture Features\end{tabular} &
  \begin{tabular}[c]{@{}c@{}}Space-Spectral Features\end{tabular} &
  \begin{tabular}[c]{@{}c@{}}Contractive Learning\end{tabular} &
  \begin{tabular}[c]{@{}c@{}}Attention Fusion \end{tabular} &
  InP &
  PaU &
  Hou &
  XuZ \\
\hline
1 & $\checkmark$ & \ding{53} & \ding{53} & \ding{53} & 0.6961 & 0.8479 & 0.6096 & 0.7678 \\
2 & \ding{53} & $\checkmark$ & \ding{53} & \ding{53} & 0.8483 & 0.8442 & 0.6238 & 0.7834 \\
3 & $\checkmark$ & $\checkmark$ & \ding{53} & \ding{53} & 0.9251 & 0.8727 & 0.8094 & 0.8895 \\
4 & $\checkmark$ & $\checkmark$ & $\checkmark$ & \ding{53} & 0.9541 & 0.9103 & 0.8545 & 0.9504 \\
5 & $\checkmark$ & $\checkmark$ & $\checkmark$ & $\checkmark$ & 0.9761 & 0.9669 & 0.8721 & 0.9765 \\
\hline
\end{tabular}
}
\end{table*}

\subsection{Ablation Experiments}
We conducted various ablation experiments on each module in the network to consider the contributions of the different modules to the rationality of the network structure. Table \ref{tb:ablation} presents the quantitative results of the model on the Indian Pines, Pavia University, Houston2013, and Xu Zhou datasets, with the results being the average of 10 experiments.

\subsubsection{Ablation experiment for multi-view:} Although we have already concluded from the previous results that multi-view complementary information can improve clustering accuracy, we demonstrated this more intuitively by comparing the results of a single spectral view, textural view, and multi-view (Table \ref{tb:ablation}). The results of a single view are not satisfactory, whereas the fusion of multi-view information can significantly improve OA accuracy. Although the individual textural view does not surpass the spatial-spectral view in terms of accuracy, the multi-view results show improvements of 7.68$\%$, 2.85$\%$, 18.56$\%$, and 10.61$\%$ on the Indian Pines, Pavia University, Houston-2013, and Xu Zhou datasets, respectively. This indicates that textural features can complement the missing information in the spatial-spectral view; using textural features, the potential spatial information of land cover can be better exploited. The combination of both leads to an improved clustering accuracy.

\subsubsection{Ablation study on contrastive learning:} Table \ref{tb:ablation} compares the clustering accuracies with and without contrastive learning loss (Cases 3 and 4). We observed that for each dataset, the model with contrastive learning consistently outperformed the simplified model without contrastive learning, achieving higher OA, NMI, and kappa accuracy. After adding contrastive learning, the model improved by 2.9$\%$, 3.76$\%$, 4.51$\%$, and 6.09$\%$ for the four datasets. This confirms that contrastive loss significantly improves clustering performance.

\subsubsection{Ablation study on adaptive affinity matrix:} In Table \ref{tb:ablation}, Case 5 represents a simple variant of our model, which only averages the affinity matrices instead of using the attention-based fusion module for spectral clustering. A comparison of Cases 4 and 5 highlights the advantages of the attention-mechanism-based adaptive fusion module. It improved the performance on the four datasets by 2.2$\%$, 5.66$\%$, 1.76$\%$, and 2.61$\%$. This indicates that effectively integrating information from multi-view feature spaces can lead to more efficient HSI clustering.

We also conducted a visual comparison of the learned affinity matrices on the Indian Pines dataset; the visualization results are presented in Figure \ref{fig:affinity}. As can be seen from the three subfigures, the affinity matrix learned by CMSCGC is not only sparse, but also exhibits a more pronounced block diagonal structure, indicating that our model is capable of better identifying intra-cluster relationships. This can be attributed to the adaptive utilization of textural and spectral information, which leads to accurate HSI classification.

\section{CONCLUSION} \label{sec:CONCLUSION}
In this paper, we develop a novel multi-view subspace clustering network called CMSCGC for HSI clustering. Specifically, we first extract textural and spectral information from HSIs to construct a multi-view graph structure. We then use GCNs to aggregate neighborhood information and employ contrastive learning to learn the consistency and complementary information of the multi-view data. The extracted robust features are then fed into a self-expression network to learn the affinity matrices. Finally, considering the varying contributions of different views to the clustering task, we adopt an attention-based strategy to combine the learning of the two branches and obtain more discriminative affinity matrices. We test our proposed model on four common hyperspectral datasets and compared it with state-of-the-art models. The experimental results show that CMSCGC achieved the best clustering performance. Comprehensive ablation studies are also conducted to verify the effectiveness of the designed modules.

The self-expression layer makes it difficult for the proposed model to be trained on large-scale datasets. Although the min-batch method can alleviate computational pressure, it is challenging to fully exploit the capabilities of GCNs. In the future, we will attempt to explore the potential of CMSCGC using large-scale datasets.

\section{ACKNOWLEDGMENT}
Computation of this study was performed by the High-performance GPU Server (TX321203) Computing Centre of the National Education Field Equipment Renewal and Renovation Loan Financial Subsidy Project of China University of Geosciences, Wuhan.

\bibliographystyle{IEEEtran}
\bibliography{IEEEfull,main}

\end{document}